\documentclass{article}

\usepackage[preprint]{neurips_2024}

\usepackage[utf8]{inputenc} % allow utf-8 input
\usepackage[T1]{fontenc}    % use 8-bit T1 fonts
\usepackage{hyperref}       % hyperlinks
\usepackage{url}            % simple URL typesetting
\usepackage{booktabs}       % professional-quality tables
\usepackage{amsfonts}       % blackboard math symbols
\usepackage{nicefrac}       % compact symbols for 1/2, etc.
\usepackage{microtype}      % microtypography
\usepackage{xcolor}         % colors
\usepackage{amsmath}
\usepackage{pifont}
\usepackage{multirow}
\usepackage{graphicx}
\usepackage{xspace}
\usepackage{lipsum} %For dummy text to test
\usepackage{anyfontsize} %Allow font sizing

\newcommand\our{\text{BitNet b1.58 2B4T}}
\newcommand{\cmark}{\ding{51}}%
\newcommand{\xmark}{\ding{55}}%
\newcommand{\huggingface}{\raisebox{-1.5pt}{\includegraphics[height=1.05em]{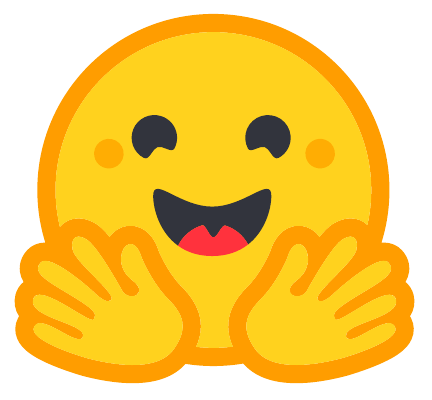}}\xspace}
\newcommand{\github}{\raisebox{-1.5pt}{\includegraphics[height=1.05em]{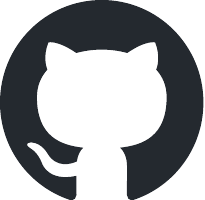}}\xspace}

\title{BitNet b1.58 2B4T Technical Report}

\vspace{-5.0cm}
\author{
Shuming Ma\thanks{~Equal contribution. $\diamond$ Corresponding author. S. Ma, S. Huang, X. Zhang, T. Song, Y. Xia and F. Wei are with Microsoft Research. H. Wang is with University of Chinese Academy of Sciences. Y. Hu is with Tsinghua University.}~~~~Hongyu Wang\footnotemark[1]~~~~Shaohan Huang~~~~Xingxing Zhang \\
\bf Ying Hu~~~~Ting Song~~~~Yan Xia~~~~Furu Wei$^{\diamond}$ \\
{\href{https://aka.ms/GeneralAI}{https://aka.ms/GeneralAI}}
\vspace{-0.2cm}
\\}

\begin{document}

\maketitle

\vspace{-0.6cm}
\begin{abstract}
  We introduce BitNet b1.58 2B4T, the first open-source, native 1-bit Large Language Model (LLM) at the 2-billion parameter scale. Trained on a corpus of 4 trillion tokens, the model has been rigorously evaluated across  benchmarks covering language understanding, mathematical reasoning, coding proficiency, and conversational ability. Our results demonstrate that BitNet b1.58 2B4T achieves performance on par with leading open-weight, full-precision LLMs of similar size, while offering significant advantages in computational efficiency, including substantially reduced memory footprint, energy consumption, and decoding latency. To facilitate further research and adoption, the model weights are released via Hugging Face along with open-source inference implementations for both GPU and CPU architectures.
\end{abstract}
 \vspace{-0.2cm}

{\small % Start scope and set size to \small
\hspace*{1.3cm}\huggingface\ \textbf{\our{} (1.58-bit)}:\ {\href{https://huggingface.co/microsoft/bitnet-b1.58-2B-4T}{\texttt{bitnet-b1.58-2B-4T}}}\\
\hspace*{1.8cm}\emph{The packed weight of \our{}, used for inference only}

\hspace*{1.3cm}\huggingface\ \textbf{\our{} (bf16)}:\ {\href{https://huggingface.co/microsoft/bitnet-b1.58-2B-4T-bf16}{\texttt{bitnet-b1.58-2B-4T-bf16}}}\\
\hspace*{1.8cm}\emph{The master weight of \our{}, used for training only}

\hspace*{1.3cm}\huggingface\ \textbf{\our{} (gguf)}:\ {\href{https://huggingface.co/microsoft/bitnet-b1.58-2B-4T-gguf}{\texttt{bitnet-b1.58-2B-4T-gguf}}}\\
\hspace*{1.8cm}\emph{The GGUF format of \our{}, used for bitnet.cpp}

\hspace*{1.3cm}\github\ \textbf{\our{} Code}:\ {\href{https://github.com/microsoft/BitNet}{\texttt{bitnet.cpp}}}
\hspace*{0.6cm}\ \textbf{Demo}:\ {\href{https://aka.ms/bitnet-demo}{\texttt{aka.ms/bitnet-demo}}}\\
} % End scope for \small

\vspace{-0.5cm}
\begin{figure}[htbp]
  \centering
  \includegraphics[width=0.85\textwidth]{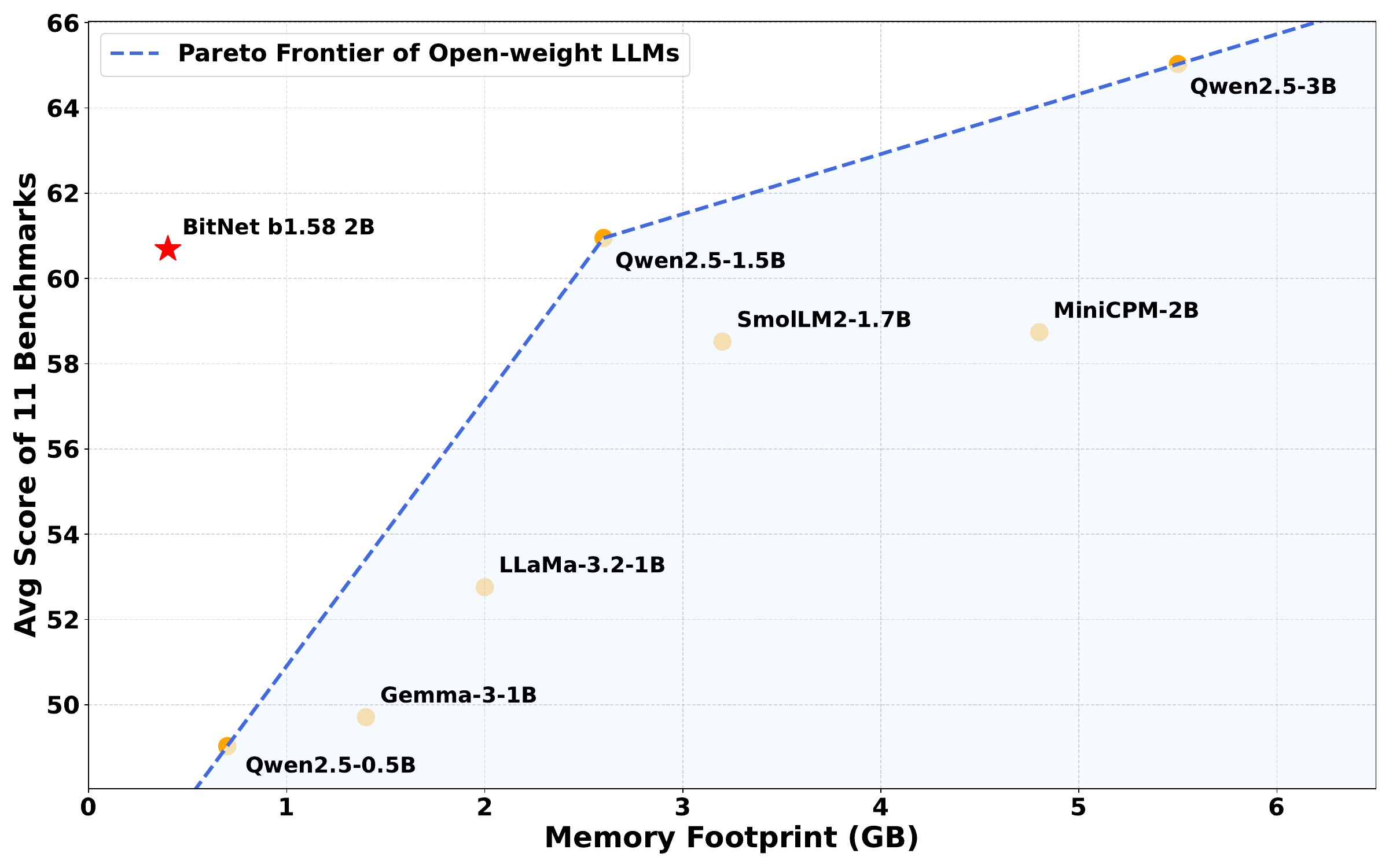} 
  \vspace{-0.2cm}
  \caption{\our{} advances the Pareto frontier defined by leading open-weight LLMs under 3B parameters in terms of performance versus memory, demonstrating superior efficiency.}
  \label{fig:frontier}
\end{figure}

\newpage

\section{Introduction}

Open-source large language models (LLMs) have become pivotal in democratizing access to advanced AI capabilities, fostering innovation, and enabling research across diverse fields such as natural language processing, code generation, and vision computing~\citep{llama3, qwen2-5, qwen2-5vl}. Their public availability allows for widespread experimentation and adaptation. However, a significant barrier hinders their broader adoption: the substantial computational resources required for deployment and inference. State-of-the-art open LLMs typically require large memory footprints, consume considerable energy, and exhibit notable inference latency, rendering them impractical for many edge devices, resource-constrained environments, and real-time applications.

1-bit LLMs, representing an extreme yet promising form of model quantization where weights and potentially activations are constrained to binary \{-1, +1\} or ternary \{-1, 0, +1\}, offer a compelling solution to the efficiency challenges. By drastically reducing the memory required to store weights and enabling highly efficient bitwise computations, they have the potential to significantly lower deployment costs, reduce energy consumption, and accelerate inference speeds. While prior work has explored 1-bit models, existing open efforts often fall into two categories: 1) post-training quantization (PTQ) methods applied to pre-trained full-precision models, which can lead to significant performance degradation~\citep{onebit, falcon3}, or 2) native 1-bit models (trained from scratch with 1-bit weights) that have been developed at relatively smaller scales (e.g., 
OLMo-Bitnet-1B\footnote{\url{https://huggingface.co/NousResearch/OLMo-Bitnet-1B}}]) and may not yet match the capabilities of larger, full-precision counterparts. This performance gap has limited the practical impact of 1-bit LLMs thus far.

To bridge this gap between efficiency and performance, we introduce \our{}, the first open-source, native 1-bit LLM trained at scale. This model, comprising 2 billion parameters, was trained from scratch on a substantial dataset of 4 trillion tokens, leveraging architectural and training innovations specific to the 1-bit paradigm. \textbf{The core contribution of this work is to demonstrate that a native 1-bit LLM, when trained effectively at scale, can achieve performance comparable to leading open-weight, full-precision models of similar size across a wide range of tasks.}

This technical report details the development and evaluation of \our{}. We describe the architecture and training methodology, and then present comprehensive evaluation results on standard benchmarks assessing language understanding, mathematical reasoning, coding proficiency, and multi-turn conversational abilities. Our findings confirm its strong performance relative to established full-precision baselines, coupled with significant advantages in efficiency. Finally, we announce the public release of the \our{} model weights via Hugging Face and provide open-source inference code optimized for both GPU and CPU execution, aiming to facilitate further research and the practical deployment of highly efficient LLMs.

\section{Architecture}

The architecture of \our{} is derived from the standard Transformer model~\citep{transformer}, incorporating significant modifications based on the BitNet framework~\citep{bitnet, bitnet-b1-58}. The model is trained entirely from scratch.

The core architectural innovation lies in replacing the standard full-precision linear layers (\emph{torch.nn.Linear}) with custom \emph{BitLinear} layers. This constitutes the foundation of the BitNet approach. Within these \emph{BitLinear} layers:
\begin{itemize}
    \item \textbf{Weight Quantization:} Model weights are quantized to 1.58 bits during the forward pass. This is achieved using an absolute mean (absmean) quantization scheme, which maps weights to ternary values $\{-1, 0, +1\}$. This drastically reduces the model size and enables efficient mathematical operations.
    \item \textbf{Activation Quantization:} Activations flowing through the linear projection are quantized to 8-bit integers. This employs an absolute maximum (absmax) quantization strategy, applied per-token.
    \item \textbf{Normalization:} We incorporate \texttt{subln} normalization \citep{magneto} to further enhance training stability, which can be particularly beneficial in quantized training regimes.
\end{itemize}

Beyond the \emph{BitLinear} layers, several established LLM techniques are integrated to enhance performance and stability:
\begin{itemize}
    \item \textbf{Activation Function (FFN):} Within the feed-forward network (FFN) sub-layers, instead of the commonly used SwiGLU activation~\citep{swiglu}, \our{} employs squared ReLU ($\text{ReLU}^2$). This choice is motivated by its potential to improve model sparsity and computational characteristics within the 1-bit context~\citep{bitnet-a4-8, qsparse}.
    \item \textbf{Positional Embeddings:} Rotary Position Embeddings (RoPE)~\citep{rope} are used to inject positional information, a standard practice in modern high-performance LLMs.
    \item \textbf{Bias Removal:} Consistent with architectures like LLaMA, all bias terms are removed from the linear layers and normalization layers throughout the network, reducing parameter count and potentially simplifying quantization.
\end{itemize}

For tokenization, we adopt the tokenizer developed for LLaMA 3 \citep{llama3}. This tokenizer implements a byte-level Byte-Pair Encoding (BPE) scheme with a vocabulary size of 128,256 tokens. This choice ensures robust handling of diverse text and code, and its widespread adoption facilitates straightforward integration with existing open-source tooling and ecosystems.

\section{Training}

The training process for \our{} involved three distinct phases: large-scale pre-training followed by supervised fine-tuning (SFT) and direct preference optimization (DPO). While advanced techniques like Proximal Policy Optimization (PPO) or Group Relative Policy Optimization (GRPO) can further enhance capabilities such as mathematics and chain-of-thought reasoning \citep{ppo, grpo}, the current version of \our{} relies solely on pre-training, SFT, and DPO. The exploration of reinforcement learning methods remains a direction for future work.

\subsection{Pre-training}

The pre-training phase aimed to imbue the model with broad world knowledge and foundational language capabilities. We adapted general training strategies from established LLM practices \citep{llama3}, with specific adjustments tailored for the 1-bit architecture.

\subsubsection{Learning Rate Schedule}
A two-stage learning rate schedule was employed.
\begin{enumerate}
    \item \textbf{Stage 1 (High Learning Rate):} The initial phase utilized a standard cosine decay schedule but commenced with a relatively high peak learning rate. This decision was informed by the observation that 1-bit models often exhibit greater training stability compared to their full-precision counterparts, allowing for more aggressive initial learning steps.
    \item \textbf{Stage 2 (Cooldown):} Approximately midway through the planned training token count, the learning rate was abruptly decayed and subsequently maintained via a cosine schedule with a significantly lower peak value. This "cooldown" phase allows the model to refine its representations on higher-quality data (see Section~\ref{sec:pretraining_data}).
\end{enumerate}

\subsubsection{Weight Decay Schedule}
Complementing the learning rate adjustments, a two-stage weight decay strategy was implemented.
\begin{enumerate}
    \item \textbf{Stage 1:} During the first training stage, weight decay followed a cosine schedule, reaching a peak value of $0.1$. This regularization helps prevent overfitting during the initial high-learning-rate phase.
    \item \textbf{Stage 2:} In the second stage, weight decay was effectively disabled (set to zero). This allows the model parameters to settle into finer-grained optima guided by the lower learning rate and curated data.
\end{enumerate}

\subsubsection{Pre-training Data} \label{sec:pretraining_data}
The pre-training corpus comprised a mixture of publicly available text and code datasets, including large web crawls like DCLM~\citep{datacamp-lm} and educational web pages like FineWeb-EDU~\citep{fineweb-edu}. To enhance mathematical reasoning abilities, we also incorporated synthetically generated mathematical data. The data presentation strategy aligned with the two-stage training: the bulk of general web data was processed during Stage 1, while higher-quality curated datasets were emphasized during the Stage 2 cooldown phase, coinciding with the reduced learning rate.

\subsection{Supervised Fine-tuning (SFT)}

Following pre-training, the model underwent supervised fine-tuning (SFT) to enhance its instruction-following capabilities and improve its performance in conversational interaction formats.

\subsubsection{SFT Data}
The SFT phase utilized a diverse collection of publicly available instruction-following and conversational datasets. These included, but were not limited to, WildChat~\citep{wildchat}, LMSYS-Chat-1M~\citep{lmsys-chat}, WizardLM Evol-Instruct~\citep{wizardlm}, and SlimOrca~\citep{SlimOrca}. To further bolster specific capabilities, particularly in reasoning and complex instruction adherence, we supplemented these with synthetic datasets generated using methodologies like GLAN~\citep{glan} and MathScale~\citep{mathscale}.

\subsubsection{Chat Template}
For conversational tasks during SFT and inference, the following chat template structure was employed:
\begin{verbatim}
<|begin_of_text|>System: {system_message}<|eot_id|>
User: {user_message_1}<|eot_id|>
Assistant: {assistant_message_1}<|eot_id|>
User: {user_message_2}<|eot_id|>
Assistant: {assistant_message_2}<|eot_id|>...
\end{verbatim}

\subsubsection{Optimization Details}
Several optimization choices were key during SFT:
\begin{itemize}
    \item \textbf{Loss Aggregation:} Instead of averaging the cross-entropy loss across tokens within a batch (mean reduction), we employed summation. Empirically, we observed that summing the losses led to improved convergence and better final performance for this model.
    \item \textbf{Hyperparameter Tuning:} Careful tuning of the learning rate and the number of training epochs was performed. Consistent with our pre-training findings, the 1-bit model benefited from a relatively larger learning rate during SFT compared to typical full-precision model fine-tuning. Furthermore, achieving optimal convergence required extending the fine-tuning duration over a larger number of epochs than full-precision models of similar size.
\end{itemize}

\subsection{Direct Preference Optimization (DPO)}

To further align the model's behavior with human preferences regarding helpfulness and safety, we applied Direct Preference Optimization (DPO)~\citep{dpo} following the SFT phase. DPO offers an efficient alternative to traditional RLHF by directly optimizing the language model using preference data, thereby circumventing the need to train a separate reward model. This DPO stage served to refine the model's conversational prowess and overall alignment with desired interaction patterns in practical use cases.

\subsubsection{Training Data}
The preference dataset used for DPO training was constructed from a combination of publicly available resources recognized for capturing diverse human judgments on model outputs. Specifically, we utilized UltraFeedback~\citep{ultrafeedback} and MagPie~\citep{magpie}.
The aggregation of these datasets provided a robust and multifaceted preference signal, guiding the model towards generating responses more aligned with human expectations.

\subsubsection{Training Details}
The DPO training phase was conducted for 2 epochs. We employed a learning rate of $2 \times 10^{-7}$ and set the DPO beta parameter, which controls the divergence from the reference policy, to 0.1. To enhance training efficiency during this phase, we integrated optimized kernels from the \emph{Liger Kernel} library~\citep{liger}. Qualitatively, our observations indicate that the DPO process effectively steered the model towards preferred response styles without inducing significant degradation in the core capabilities established during pre-training and SFT.

\begin{table}[t]
    \centering
    \footnotesize
    \setlength{\tabcolsep}{4.5pt}
    \begin{tabular}{c | c c c c c | c}
    \toprule
    \multirow{2}{*}{\centering \textbf{Benchmark {\tiny (Metric)}}} & \textbf{LLaMA 3.2} & \textbf{Gemma-3} & \textbf{Qwen2.5} & \textbf{SmolLM2} & \textbf{MiniCPM} & \textbf{BitNet b1.58}  \\
    & \textbf{1B} & \textbf{1B} & \textbf{1.5B} & \textbf{1.7B} & \textbf{2B} & \textbf{2B} \\
    \midrule
    Memory & \multirow{2}{*}{2GB} & \multirow{2}{*}{1.4GB} & \multirow{2}{*}{2.6GB} & \multirow{2}{*}{3.2GB} & \multirow{2}{*}{4.8GB} & \multirow{2}{*}{\bf 0.4GB} \\
    {\tiny (Non-emb)} & & & & & & \\
    Latency & \multirow{2}{*}{48ms} & \multirow{2}{*}{41ms} & \multirow{2}{*}{65ms} & \multirow{2}{*}{67ms} & \multirow{2}{*}{124ms} & \multirow{2}{*}{\bf 29ms} \\
    {\tiny (CPU; TPOT)} & & & & & & \\
    Energy & \multirow{2}{*}{0.258J} & \multirow{2}{*}{0.186J} & \multirow{2}{*}{0.347J} & \multirow{2}{*}{0.425J} & \multirow{2}{*}{0.649J} & \multirow{2}{*}{\bf 0.028J} \\
    {\tiny (Estimated)} & & & & & & \\
    Training Tokens & \multirow{2}{*}{9T} & \multirow{2}{*}{2T} & \multirow{2}{*}{18T} & \multirow{2}{*}{11T} & \multirow{2}{*}{1.1T} & \multirow{2}{*}{4T}\\
    {\tiny (Pre-training)} & {\tiny (pruning \& distillation)} & {\tiny (distillation)} & & & & \\
    \midrule
    ARC-Challange   & \multirow{2}{*}{37.80} & \multirow{2}{*}{38.40} & \multirow{2}{*}{46.67} & \multirow{2}{*}{43.52} & \multirow{2}{*}{44.80} & \multirow{2}{*}{\bf 49.91} \\
    {\tiny (0-shot; Acc,norm)} & & & & & & \\
    ARC-Easy   & \multirow{2}{*}{63.17} & \multirow{2}{*}{63.13} & \multirow{2}{*}{\bf 76.01} & \multirow{2}{*}{62.92} & \multirow{2}{*}{72.14} & \multirow{2}{*}{74.79} \\
    {\tiny (0-shot; Acc,norm)} & & & & & & \\
    OpenbookQA  & \multirow{2}{*}{34.80} & \multirow{2}{*}{38.80} & \multirow{2}{*}{40.80} & \multirow{2}{*}{\bf 46.00} & \multirow{2}{*}{40.20} & \multirow{2}{*}{41.60} \\
    {\tiny (0-shot; Acc,norm)} & & & & & & \\
    BoolQ  & \multirow{2}{*}{64.65} & \multirow{2}{*}{74.22} & \multirow{2}{*}{78.04} & \multirow{2}{*}{75.78} & \multirow{2}{*}{\bf 80.67} & \multirow{2}{*}{80.18}\\
    {\tiny (0-shot; Acc)} & & & & & & \\
    HellaSwag  & \multirow{2}{*}{60.80} & \multirow{2}{*}{57.69} & \multirow{2}{*}{68.28} & \multirow{2}{*}{\bf 71.71} & \multirow{2}{*}{70.81} & \multirow{2}{*}{68.44} \\
    {\tiny (0-shot; Acc,norm)} & & & & & & \\
    PIQA  & \multirow{2}{*}{74.21} & \multirow{2}{*}{71.93} & \multirow{2}{*}{76.12} & \multirow{2}{*}{76.12} & \multirow{2}{*}{76.66} & \multirow{2}{*}{\bf 77.09} \\
    {\tiny (0-shot; Acc,norm)} & & & & & & \\
    WinoGrande & \multirow{2}{*}{59.51} & \multirow{2}{*}{58.48} & \multirow{2}{*}{62.83} & \multirow{2}{*}{68.98} & \multirow{2}{*}{61.80} & \multirow{2}{*}{\bf 71.90}\\
    {\tiny (0-shot; Acc)} & & & & & & \\
    CommonsenseQA & \multirow{2}{*}{58.48} & \multirow{2}{*}{42.10} & \multirow{2}{*}{\bf 76.41} & \multirow{2}{*}{63.55} & \multirow{2}{*}{71.74} & \multirow{2}{*}{71.58}\\
    {\tiny (10-shot; Acc)} & & & & & & \\
    TruthfulQA & \multirow{2}{*}{43.80} & \multirow{2}{*}{38.66} & \multirow{2}{*}{\bf 46.67} & \multirow{2}{*}{39.90} & \multirow{2}{*}{41.41} & \multirow{2}{*}{45.31} \\
    {\tiny (10-shot; MC2)} & & & & & & \\
    TriviaQA & \multirow{2}{*}{37.60} & \multirow{2}{*}{23.49} & \multirow{2}{*}{38.37} & \multirow{2}{*}{\bf 45.97} & \multirow{2}{*}{34.13} & \multirow{2}{*}{33.57} \\
    {\tiny (5-shot; EM)} & & & & & & \\
    MMLU & \multirow{2}{*}{45.58} & \multirow{2}{*}{39.91} & \multirow{2}{*}{\bf 60.25} & \multirow{2}{*}{49.24} & \multirow{2}{*}{51.82} & \multirow{2}{*}{53.17} \\
    {\tiny (5-shot; Acc)} & & & & & & \\
    \midrule
    % HumanEval & \\
    % {\tiny (0-shot; Pass@1)} & \\
    HumanEval+ & \multirow{2}{*}{31.10} & \multirow{2}{*}{37.20} & \multirow{2}{*}{\bf 50.60} & \multirow{2}{*}{28.00} & \multirow{2}{*}{43.90} & \multirow{2}{*}{38.40} \\
    {\tiny (0-shot; Pass@1)} & & & & & & \\
    % \midrule
    GSM8K & \multirow{2}{*}{38.21} & \multirow{2}{*}{31.16} & \multirow{2}{*}{56.79} & \multirow{2}{*}{45.11} & \multirow{2}{*}{4.40} & \multirow{2}{*}{\bf 58.38}\\
    {\tiny (4-shot; EM)} & & & & & & \\
    MATH-500 & \multirow{2}{*}{23.00} & \multirow{2}{*}{42.00} & \multirow{2}{*}{\bf 53.00} & \multirow{2}{*}{17.60} & \multirow{2}{*}{14.80} & \multirow{2}{*}{43.40}\\
    {\tiny (0-shot; EM)} & & & & & & \\
    \midrule
    IFEval & \multirow{2}{*}{62.71} & \multirow{2}{*}{\bf 66.67} & \multirow{2}{*}{50.12} & \multirow{2}{*}{57.91} & \multirow{2}{*}{36.81} & \multirow{2}{*}{53.48} \\
    {\tiny (0-shot; Instruct-Strict)} & & & & & & \\
    MT-bench & \multirow{2}{*}{5.43} & \multirow{2}{*}{6.40} & \multirow{2}{*}{6.12} & \multirow{2}{*}{5.50} & \multirow{2}{*}{\bf 6.57} & \multirow{2}{*}{5.85} \\
    {\tiny (0-shot; Average)} & & & & & & \\
    \midrule
    Average & 44.90 & 43.74 & \bf 55.23 & 48.70 & 42.05 & 54.19 \\
    \bottomrule
    \end{tabular}
    \vspace{0.2cm}
    \caption{Comparison of \our{} with leading open-weight full-precision LLMs of similar size (1B-2B parameters) on efficiency metrics and performance across a wide range of benchmarks. All models compared are instruction-tuned versions.
    }
    \label{tab:main}
\end{table}

\section{Evaluation}

\begin{table}[t]
    \centering
    % \footnotesize
    \setlength{\tabcolsep}{5pt}
    \begin{tabular}{c | c c c | c}
    \toprule
    \multirow{2}{*}{\centering \textbf{Benchmark {\tiny (Metric)}}} & \multicolumn{3}{c|}{\textbf{Qwen2.5}} & \textbf{BitNet b1.58}  \\
    & \textbf{1.5B-bf16} & \textbf{1.5B-GPTQ-int4} & \textbf{1.5B-AWQ-int4} & \textbf{2B} \\
    \midrule
    Memory & \multirow{2}{*}{2.6GB} & \multirow{2}{*}{0.7GB} & \multirow{2}{*}{0.7GB} & \multirow{2}{*}{0.4GB} \\
    {\tiny (Non-emb)} & & & & \\
    Activation & bf16 & bf16 & bf16 & int8 \\
    \midrule
    MMLU & \multirow{2}{*}{\bf 60.25} & \multirow{2}{*}{58.06} & \multirow{2}{*}{57.43} & \multirow{2}{*}{53.17} \\
    {\tiny (5-shot; Acc)} & & & & \\
    GSM8K & \multirow{2}{*}{56.79} & \multirow{2}{*}{50.57} & \multirow{2}{*}{50.64} & \multirow{2}{*}{\bf 58.38}\\
    {\tiny (4-shot; EM)} & & & & \\
    IFEval & \multirow{2}{*}{50.12} & \multirow{2}{*}{47.84} & \multirow{2}{*}{45.44} & \multirow{2}{*}{\bf 53.48} \\
    {\tiny (0-shot; Instruct-Strict)} & & & & \\
    \midrule
    Average & \bf 55.72 & 52.15 & 51.17 & 55.01 \\
    \bottomrule
    \end{tabular}
    \vspace{0.2cm}
    \caption{Comparison of BitNet b1.58 (2B) against Qwen2.5 1.5B in its original bf16 precision and after INT4 post-training quantization (GPTQ and AWQ). All models shown are based on instruction-tuned checkpoints.
    }
    \label{tab:ptq}
\end{table}

\begin{table}[!ht]
    \centering
    \footnotesize
    \setlength{\tabcolsep}{4.5pt}
    \begin{tabular}{c | c c c c | c}
    \toprule
    \multirow{2}{*}{\centering \textbf{Benchmark {\tiny (Metric)}}} & \textbf{Bonsai} & \textbf{OLMo-Bitnet} & \textbf{Falcon3-1.58bit} & \textbf{Llama3-8B-1.58} & \textbf{BitNet b1.58}  \\
    & \textbf{0.5B} & \textbf{1B} & \textbf{7B} & \textbf{8B} & \textbf{2B} \\
    \midrule
    Native 1-bit & \cmark & \cmark & \xmark & \xmark & \cmark \\
    \midrule
    ARC-Challange   & \multirow{2}{*}{33.19} & \multirow{2}{*}{26.54} & \multirow{2}{*}{37.80} & \multirow{2}{*}{43.69} & \multirow{2}{*}{\bf 49.91} \\
    {\tiny (0-shot; Acc,norm)} & & & & & \\
    ARC-Easy   & \multirow{2}{*}{58.25} & \multirow{2}{*}{25.38} & \multirow{2}{*}{65.03} & \multirow{2}{*}{70.71} & \multirow{2}{*}{\bf 74.79} \\
    {\tiny (0-shot; Acc,norm)} & & & & & \\
    OpenbookQA  & \multirow{2}{*}{33.60} & \multirow{2}{*}{28.20} & \multirow{2}{*}{38.20} & \multirow{2}{*}{37.20} & \multirow{2}{*}{\bf 41.60} \\
    {\tiny (0-shot; Acc,norm)} & & & & & \\
    BoolQ  & \multirow{2}{*}{58.44} & \multirow{2}{*}{52.48} & \multirow{2}{*}{72.14} & \multirow{2}{*}{68.38} & \multirow{2}{*}{\bf 80.18} \\
    {\tiny (0-shot; Acc)} & & & & & \\
    HellaSwag  & \multirow{2}{*}{48.01} & \multirow{2}{*}{25.88} & \multirow{2}{*}{59.46} & \multirow{2}{*}{\bf 68.56} & \multirow{2}{*}{ 68.44} \\
    {\tiny (0-shot; Acc,norm)} & & & & & \\
    PIQA  & \multirow{2}{*}{70.02} & \multirow{2}{*}{50.49} & \multirow{2}{*}{72.36} & \multirow{2}{*}{75.30} & \multirow{2}{*}{\bf 77.09} \\
    {\tiny (0-shot; Acc,norm)} & & & & & \\
    WinoGrande & \multirow{2}{*}{54.46} & \multirow{2}{*}{51.54} & \multirow{2}{*}{60.14} & \multirow{2}{*}{60.93} & \multirow{2}{*}{\bf 71.90} \\
    {\tiny (0-shot; Acc)} & & & & & \\
    CommonsenseQA  & \multirow{2}{*}{18.43} & \multirow{2}{*}{19.49} & \multirow{2}{*}{67.08} & \multirow{2}{*}{28.50} & \multirow{2}{*}{\bf 71.58} \\
    {\tiny (10-shot; Acc)} & & & & & \\
    TruthfulQA & \multirow{2}{*}{40.65} & \multirow{2}{*}{\bf 49.05} & \multirow{2}{*}{43.29} & \multirow{2}{*}{39.13} & \multirow{2}{*}{45.31} \\
    {\tiny (10-shot; MC2)} & & & & & \\
    TriviaQA & \multirow{2}{*}{10.84} & \multirow{2}{*}{0.00} & \multirow{2}{*}{0.00} & \multirow{2}{*}{19.82} & \multirow{2}{*}{\bf 33.57} \\
    {\tiny (5-shot; EM)} & & & & & \\
    MMLU & \multirow{2}{*}{25.74} & \multirow{2}{*}{25.47} & \multirow{2}{*}{42.79} & \multirow{2}{*}{35.04} & \multirow{2}{*}{\bf 53.17} \\
    {\tiny (5-shot; Acc)} & & & & & \\
    \midrule
    Average & 41.06 & 32.22 & 50.76 & 49.75 & \bf 60.68 \\
    \bottomrule
    \end{tabular}
    \vspace{0.2cm}
    \caption{Performance comparison of \our{} against other open-weight 1-bit models. This includes natively trained 1-bit models (Bonsai-0.5B, OLMo-Bitnet-1B) and larger models post-training quantized to 1.58-bit (Falcon3-1.58bit-7B, Llama3-8B-1.58).
    }
    \label{tab:onebit}
\end{table}

We measure performance on a wide variety of benchmarks classified as follows:

\begin{itemize}
    \item \textbf{Language understanding and reasoning:} ARC-Easy \citep{arc}, ARC-Challenge \citep{arc}, HellaSwag \citep{hellaswag}, WinoGrande \citep{winoGrande}, PIQA \citep{piqa}, OpenbookQA \citep{openbookqa}, and CommonsenseQA \citep{commonsenseqa}
    \item \textbf{World knowledge:} TruthfulQA~\citep{truthfulqa} and MMLU~\citep{mmlu}
    \item \textbf{Reading comprehension:} TriviaQA~\citep{triviaqa} and BoolQ~\citep{boolq}
    \item \textbf{Math and code:} GSM8K \citep{gsm8k},  MATH-500 \citep{math} and HumanEval+ \citep{humaneval-plus}
    \item \textbf{Instruction following and conversation:} IFEval \citep{ifeval} and MT-bench \citep{mtbench}
\end{itemize}

We compare \our{} with leading open-weight full precision LLMs of similar size, including LLaMA 3.2 1B~\citep{llama3}, Gemma-3 1B~\citep{gemma3}, Qwen2.5 1.5B~\citep{qwen2-5}, SmolLM2 1.7B~\citep{smollm2} and MiniCPM 2B~\citep{minicpm}. All models are instruction-tuned versions.
We re-run all benchmarks with a public evaluation pipeline for a fair comparison. More evaluation details are available at the appendix. The main results are presented in Table~\ref{tab:main}.

\subsection{Main Results}

As shown in Table 1, \our{} demonstrates remarkable resource efficiency. Its non-embedding memory footprint and estimated energy consumption~\citep{energycost,pokebnn} during decoding are substantially lower compared to all the full-precision models evaluated, highlighting a significant advantage in operational cost and deployability on resource-constrained devices.

In terms of task performance, \our{} proves highly competitive. It achieves the best results among the compared models on several benchmarks spanning reasoning, knowledge, and math capabilities. On other benchmarks, its performance is closely comparable to the top-performing full-precision models. While some full-precision models show slight advantages on specific tasks or the overall average, \our{} delivers strong performance across the board. The results indicate that \our{} achieves capabilities nearly on par with leading models in its size class while offering dramatically improved efficiency.

\subsection{Comparison with Post-training Quantized Models}

We further investigate the efficiency-performance trade-off by comparing \our{} against post-training quantized (PTQ) versions of a leading competitor, Qwen2.5 1.5B, using standard INT4 methods (GPTQ and AWQ). The results are summarized in Table~\ref{tab:ptq}.

While INT4 quantization successfully reduces the memory footprint of the full-precision model, \our{} achieves an even lower memory requirement due to its native 1-bit architecture. More importantly, this superior memory efficiency does not compromise performance relative to the quantized models. Standard PTQ techniques lead to a noticeable degradation in performance compared to the original full-precision model. In contrast, \our{} maintains stronger overall performance than the INT4 quantized versions of Qwen2.5-1.5B on the evaluated benchmarks. This comparison suggests that \our{} represents a more favorable point on the efficiency-performance curve than applying conventional INT4 PTQ to existing architectures, offering better performance with lower resource usage.

\subsection{Comparison with Open-weight 1-bit Models}

Finally, we situate \our{} within the landscape of other models designed for or quantized to near 1-bit precision. We compare it against natively trained 1-bit models of smaller scale and significantly larger models post-training quantized to 1.58-bit precision. The comparative results are presented in Table~\ref{tab:onebit}.

The evaluation clearly positions \our{} as the leading model in this category. It demonstrates significantly stronger overall performance than all other compared 1-bit models, achieving the highest scores on the vast majority of benchmarks. Notably, \our{} substantially outperforms not only the smaller, natively trained 1-bit models but also the much larger models (in terms of parameter count) that were quantized to 1-bit. This highlights the effectiveness of the native training approach employed by \our{}, allowing it to set a new state-of-the-art performance level for models operating at this extreme level of quantization, even surpassing larger models subjected to post-training quantization.

\section{Inference Implementation}

Efficient inference is crucial for deploying Large Language Models, particularly for resource-constrained environments. The unique quantization scheme of \our{}, employing 1.58-bit weights and 8-bit activations (W1.58A8), necessitates specialized implementations, as standard deep learning libraries often lack optimized kernels for such mixed-precision, low-bit formats. To address this, we developed and open-sourced dedicated inference libraries for both GPU and CPU platforms. The code is publicly available at \url{https://aka.ms/bitnet}.

\subsection{GPU Inference}

Current GPU architectures and their associated software libraries (e.g., cuBLAS, PyTorch kernels) are primarily optimized for operations involving standard data types like FP16, BF16, and INT8/INT4. Native, high-performance support for the specific W1.58A8 matrix multiplication required by \our{} is generally unavailable. This limitation can hinder the realization of the theoretical efficiency gains offered by 1-bit models on existing hardware.

To enable efficient GPU inference, we developed a custom CUDA kernel specifically designed for the W1.58A8 matrix multiplication. Since ternary weights (\{-1, 0, +1\}, representing 1.58 bits) cannot be stored efficiently using standard data types, we pack multiple weight values into a single 8-bit integer (`int8') for storage in High Bandwidth Memory (HBM). Specifically, four ternary values are encoded into one `int8' value. During computation, the CUDA kernel loads the packed `int8' weights from HBM into the GPU's faster on-chip Shared Memory (SRAM). It then unpacks these values back into a representation suitable for efficient ternary computation (e.g., reconstructing the -1, 0, +1 values) immediately before performing the matrix multiplication with the 8-bit activations. This `pack-store-load-unpack-compute' strategy minimizes memory bandwidth usage while leveraging custom compute instructions.
Further implementation details and optimization strategies are elaborated in the Ladder framework~\citep{ladder}.

While our custom kernel significantly improves performance compared to naive implementations, we note that current commodity GPU architectures are not optimally designed for the 1-bit models. We believe that future hardware innovations, potentially incorporating dedicated logic for low-bit operations, will be essential to fully unlock the performance and energy efficiency potential of models like BitNet b1.58.

\subsection{CPU Inference}

To ensure broad accessibility and enable deployment on devices lacking powerful GPUs (e.g., edge devices, laptops, standard servers), we developed \emph{bitnet.cpp}. This C++ library serves as an official reference implementation for CPU inference of 1-bit LLMs, including BitNet b1.58.

\emph{bitnet.cpp} provides optimized kernels tailored for efficient execution on standard CPU architectures. The kernels are designed to operate efficiently with the model's specific quantization scheme, avoiding the overhead of generic quantization libraries or intricate low-level bit manipulation where possible. It processes the weight elements in a manner consistent with the BitNet b1.58 training methodology, ensuring numerical accuracy (lossless inference relative to the training procedure).

This approach delivers fast and accurate inference of 1.58-bit models directly on CPUs. More technical details and usage instructions can be found in the \emph{bitnet.cpp} repository and associated technical report~\citep{bitnetcpp}.

\section{Conclusion}

This technical report introduced \our{}, a significant step towards highly efficient yet capable Large Language Models. As the first open-source, native 1-bit LLM trained at the 2-billion parameter scale on 4 trillion tokens, our work demonstrates the viability of extreme quantization directly within the training process.

Comprehensive evaluations across benchmarks assessing language understanding, reasoning, mathematics, coding, and dialogue revealed that \our{} achieves performance comparable to state-of-the-art open-weight, full-precision models of similar size. Crucially, this performance parity is achieved with dramatically reduced computational requirements, offering substantial savings in memory footprint, energy consumption, and inference latency. To facilitate practical use and further research, we developed and released optimized inference implementations for both GPU (via custom CUDA kernels) and CPU (via the `bitnet.cpp’ library), alongside the model weights available on Hugging Face.

\our{} represents a compelling proof-of-concept that challenges the necessity of full-precision weights for achieving high performance in LLMs at scale. It opens avenues for deploying powerful language models in resource-constrained environments where previous models were prohibitive, potentially democratizing access to advanced AI capabilities.

\section{Future Directions}

While \our{} demonstrates promising results, several exciting research directions remain:

\begin{itemize}
    \item \textbf{Scaling Laws and Larger Models:} Investigating the scaling properties of native 1-bit LLMs is crucial. Future work will explore training larger models (e.g., 7B, 13B parameters and beyond) and training on even larger datasets to understand if the performance parity with full-precision models holds.

    \item \textbf{Hardware Co-Design and Optimization:} The full potential of 1-bit models is likely hindered by current hardware limitations. Continued development of highly optimized kernels for existing hardware (GPUs, CPUs, NPUs) is needed. Furthermore, co-designing future hardware accelerators specifically optimized for 1-bit computations and data movement could unlock orders-of-magnitude improvements in speed and energy efficiency.

    \item \textbf{Extended Sequence Length:} Extending the maximum sequence length of \our{} can process is crucial. This enhancement is vital for tasks demanding long-context understanding, such as summarizing lengthy documents or engaging in complex problem-solving, and is particularly critical for improving performance on \textbf{long chain-of-thought reasoning} tasks. Investigating efficient attention mechanisms suitable for low-bit models at longer sequence lengths will be key.

    \item \textbf{Multilingual Capabilities:} The current model is primarily trained on English-centric data. Extending the pre-training corpus and potentially adapting the architecture to effectively support \textbf{multiple languages} is a key direction for broader applicability.

    \item \textbf{Multimodal Integration:} Exploring the integration of 1-bit principles into \textbf{multimodal architectures} is another promising frontier. Developing efficient ways to process and fuse information from different modalities (e.g., text and images) within a low-bit framework could enable new applications.

    \item \textbf{Theoretical Understanding:} Delving deeper into the theoretical underpinnings of why 1-bit training at scale is effective remains an open area. Analyzing the learning dynamics, loss landscapes, and representational properties of these models could yield valuable insights for future development.
\end{itemize}

By pursuing these directions, we aim to further advance the capability and efficiency of 1-bit LLMs, paving the way for more sustainable and accessible artificial intelligence. The open-source release of \our{} and its associated tools provides a foundation for the community to build upon these efforts.

\bibliography{bitnet}
\bibliographystyle{apalike}

\appendix

\section{Open-weight Baselines}

We summarize the links to the open-weight LLMs evaluated in this work as below:

\begin{itemize}
\item  \textbf{LLaMA 3.2 1B}:\ {\href{https://huggingface.co/meta-llama/Llama-3.2-1B-Instruct}{\texttt{meta-llama/Llama-3.2-1B-Instruct}}}
\item  \textbf{Gemma-3 1B}:\ {\href{https://huggingface.co/google/gemma-3-1b-it}{\texttt{google/gemma-3-1b-it}}}
\item  \textbf{Qwen2.5 0.5B}:\ {\href{https://huggingface.co/Qwen/Qwen2.5-0.5B-Instruct}{\texttt{Qwen/Qwen2.5-0.5B-Instruct}}}
\item  \textbf{Qwen2.5 1.5B}:\ {\href{https://huggingface.co/Qwen/Qwen2.5-1.5B-Instruct}{\texttt{Qwen/Qwen2.5-1.5B-Instruct}}}
\item  \textbf{Qwen2.5 3B}:\ {\href{https://huggingface.co/Qwen/Qwen2.5-3B-Instruct}{\texttt{Qwen/Qwen2.5-3B-Instruct}}} 
\item  \textbf{SmolLM2 1.7B}:\ {\href{https://huggingface.co/HuggingFaceTB/SmolLM2-1.7B-Instruct}{\texttt{HuggingFaceTB/SmolLM2-1.7B-Instruct}}} 
\item  \textbf{MiniCPM 2B}:\ {\href{https://huggingface.co/openbmb/MiniCPM-2B-dpo-bf16}{\texttt{openbmb/MiniCPM-2B-dpo-bf16}}} 
\item  \textbf{Qwen2.5 1.5B-GPTQ-int4}:\ {\href{https://huggingface.co/Qwen/Qwen2.5-1.5B-Instruct-GPTQ-Int4}{\texttt{Qwen/Qwen2.5-1.5B-Instruct-GPTQ-Int4}}} 
\item  \textbf{Qwen2.5 1.5B-AWQ-int4}:\ {\href{https://huggingface.co/Qwen/Qwen2.5-1.5B-Instruct-AWQ}{\texttt{Qwen/Qwen2.5-1.5B-Instruct-AWQ}}} 
\item  \textbf{Bonsai 0.5B}:\ {\href{https://huggingface.co/deepgrove/Bonsai}{\texttt{deepgrove/Bonsai}}} 
\item  \textbf{OLMo-Bitnet 1B}:\ {\href{https://huggingface.co/NousResearch/OLMo-Bitnet-1B}{\texttt{NousResearch/OLMo-Bitnet-1B}}} 
\item  \textbf{Falcon3-1.58bit 7B}:\ {\href{https://huggingface.co/tiiuae/Falcon3-7B-Instruct-1.58bit}{\texttt{tiiuae/Falcon3-7B-Instruct-1.58bit}}} 
\item  \textbf{Llama3-8B-1.58 8B}:\ {\href{https://huggingface.co/HF1BitLLM/Llama3-8B-1.58-100B-tokens}{\texttt{HF1BitLLM/Llama3-8B-1.58-100B-tokens}}} 
\end{itemize}

\section{Evaluation Pipeline Details}

To ensure standardized evaluation, we employed established toolkits for different benchmark categories. Specifically:
\begin{itemize}
    \item For the HumanEval+ coding benchmark, we utilized the {\href{https://github.com/evalplus/evalplus}{\texttt{evalplus}}} toolkit.
    \item For the MATH-500 mathematical reasoning benchmark, we used a customized version of the {\href{https://github.com/ZubinGou/math-evaluation-harness}{\texttt{math-evaluation-harness}}} toolkit.
    \item For the MT-Bench conversational benchmark, evaluation was performed using the official {\href{https://github.com/lm-sys/FastChat/blob/main/fastchat/llm_judge/README.md}{\texttt{LLM Judge}}} open-source codebase.
    \item For all other benchmarks assessing language understanding, reasoning, knowledge, and comprehension, we used the standard {\href{https://github.com/EleutherAI/lm-evaluation-harness}{\texttt{lm-evaluation-harness}}} framework.
\end{itemize}

Models were prompted using a chat format for generative tasks (e.g., GSM8K, IFEval, and MT-Bench), while default settings from the respective toolkits were used for other tasks.

\begin{table*}[h]
\centering
    \begin{tabular}{lcc}
    \toprule
    Bits & ADD Energy & MUL Energy  \\
    \midrule
    FP16 & 0.16 & 0.34 \\
    INT8 & 0.007 & 0.07 \\
    \bottomrule
    \end{tabular}
    \caption{ADD and MUL energy consumption (in pJ) of different precision at 7nm process nodes.}
    \label{tab:bit_energy}
\end{table*}

For energy consumption, we utilize the energy model in~\citep{energycost,pokebnn} to estimate the arithmetic operations energy (AOE) of matrix multiplication. The sequence length is set as 512 tokens. We present the energy consumption for ADD and MUL operation at 7nm process nodes in Table~\ref{tab:bit_energy}.

To assess CPU decoding performance, latency measurements were conducted on a Surface Laptop Studio 2 system powered by a 13th Gen Intel Core i7-13800H processor. The benchmarking process utilized 8 CPU threads. Specifically, the \our{} model was tested using its \emph{bitnet.cpp} implementation, whereas other models were evaluated using the \emph{llama.cpp} framework. For each model, we generated 128 tokens and report the average latency per token for this task.

\end{document}